\documentclass[conference]{IEEEtran}
\IEEEoverridecommandlockouts
\usepackage{cite}
\usepackage[numbers,sort&compress]{natbib}

\usepackage[utf8]{inputenc}
\usepackage{amsmath,amssymb}
\usepackage{graphicx}
\usepackage{bm}
\usepackage{enumerate}
\usepackage{comment}
\usepackage{xcolor}
\usepackage{url}
\usepackage{color,soul}
\usepackage{subcaption}
\usepackage{float}

\definecolor{light-gray}{gray}{0.8}

\usepackage{amsmath,amssymb,amsfonts}
\usepackage{algorithmic}
\usepackage{graphicx}
\usepackage{textcomp}
\usepackage{xcolor}
\usepackage{subcaption}

\def\BibTeX{{\rm B\kern-.05em{\sc i\kern-.025em b}\kern-.08em
    T\kern-.1667em\lower.7ex\hbox{E}\kern-.125emX}}

\makeatletter
\newcommand{\linebreakand}{%
  \end{@IEEEauthorhalign}
  \hfill\mbox{}\par
  \mbox{}\hfill\begin{@IEEEauthorhalign}
}
\makeatother

\begin{document}

\title{Anomaly Detection and Early Warning Mechanism for Intelligent Monitoring Systems in Multi-Cloud Environments Based on LLM\\}

\author{

\small 

\begin{tabular}[t]{c@{\extracolsep{8em}}c} 

1\textsuperscript{st} Yihong Jin \textsuperscript{*}  & 2\textsuperscript{nd} Ze Yang \\
\textit{University of Illinois Urbana-Champaign} & \textit{University of Illinois Urbana-Champaign} \\
Champaign, USA & Champaign, USA \\
\textsuperscript{*}Corresponding Author: yihongj3@illinois.edu &  zeyang2@illinois.edu\\

\\

3\textsuperscript{rd} Juntian Liu & 4\textsuperscript{th} Xinhe Xu \\
\textit{Computer Science Department} & \textit{Computer Science Department} \\
\textit{University of Illinois Urbana-Champaign} & \textit{University of Illinois Urbana-Champaign} \\
Champaign, USA & Champaign, USA \\
jl203@illinois.edu & xinhexu2@illinois.edu  \\

\\
\end{tabular}
}

\maketitle

\begin{abstract}
With the rapid development of multi-cloud environments, it is increasingly important to ensure the security and reliability of intelligent monitoring systems, a goal that aligns with broader advancements in AI-aided infrastructure management such as digital twin design \cite{hao2024artificial}. In this paper, we propose an anomaly detection and early warning mechanism for intelligent monitoring system in multi-cloud environment based on Large-Scale Language Model (LLM). On the basis of the existing monitoring framework, the proposed model innovatively introduces a multi-level feature extraction method, which combines the natural language processing ability of LLM with traditional machine learning methods to enhance the accuracy of anomaly detection and improve the real-time response efficiency. By introducing the contextual understanding capabilities of LLMs, the model dynamically adapts to different cloud service providers and environments, so as to more effectively detect abnormal patterns and predict potential failures. Experimental results show that the proposed model is significantly better than the traditional anomaly detection system in terms of detection accuracy and latency, and significantly improves the resilience and active management ability of cloud infrastructure. 
\end{abstract}


\section{Introduction}

As cloud computing technologies continue to evolve, multi-cloud environments have rapidly become an essential component of enterprise IT architectures. Multi-cloud environments enable greater flexibility, reliability, and disaster recovery by pooling resources from multiple cloud service providers. Enterprises can have more flexibility to choose products and services from different cloud service providers according to their needs, avoiding vendor lock-in, thus improving operational efficiency and reducing costs \cite{yang2024hades}. 

Alongside this, the multi-cloud architecture allows companies to balance and optimize the charge using various cloud solutions for efficient resource usage. But this variety of architecture also brings with it a level of complexity that is new, particularly in terms of monitoring and management. “Traditional monitoring systems based on static thresholds and rules struggle to adapt to the dynamical nature of multi-clouds, leading to low accuracy and timeliness of anomaly detection \cite{hussain2024autonomous}. The architectures and interfaces provided by these cloud service providers are different, and traditional monitoring tools cannot properly collect and integrate the monitoring data of these platforms, which brings great challenges to the resource management, fault diagnosis, and performance optimization of enterprises with applications deployed on the above cloud platforms \cite{sadeq2022proactive}.

Anomaly detection is not just located as a method to detect system failure at a multi-cloud environment, it is an essential factor to guarantee the system safety and stability. Due to the euphoria of cloud computing technology, enterprises increasingly use cloud platforms for important applications and data, so the reliability and availability of cloud services is crucial \cite{li2024advances}. There could be many reasons behind abnormal behavior such as network attacks, resource allocation issues, inter-service communication failures, API call failures, hardware failures, etc. 

These problems not only disrupt service continuity, but they can also significantly affect a business's operating efficiency and customer experience. A traditional rule-based detection approach is not viable in this context as rules become too complex to handle with more dynamic multi-cloud architectures. While rules and thresholds need to be tightly defined, the rules have to become more dynamic and cover a broader scope in a multi-cloud environment as cloud platforms mature and the business scales \cite{yang2025research}. Hence, requiring a system that can detect anomalous behavior in the cloud in real-time and accurately, which should be capable of dynamic adaptation to various cloud environments and service providers for maintaining system stability and data security \cite{ji2025cloud, li2024nlp}.

Large-scale language models (LLMs) and generative models have advanced rapidly in the past years in the field of natural language processing \cite{beharidecision, sehanobishscalable}, and their powerful context understanding and generation capabilities, often enhanced by techniques such as Retrieval-Augmented Generation (RAG) \cite{ji-etal-2024-rag}, inspire new ideas for anomaly detection. Such a vast amount of numbers may be difficult to interpret, so LLMs can help process the surround logs, events, and condition strings available on the system to detect potential anomalies \cite{jin2025adaptive}. Conventional anomaly detection techniques were based on rules on structured data and ignore text data and log information that contains useful information.

This manual analysis can no longer meet the needs of real-time analysis and other requirements, due to the explosion of log data in the cloud environment. It can achieve in-depth analysis and identification for complex abnormal behaviors over a multi-cloud environment by utilizing LLM in the intelligent monitoring system. LLMs own strong contextual modeling capabilities allow detection of anomalies by predicting the behavior variations of events in a system over a period of time based on a combination of contextual features that represent historical events and information to capture and extract hidden anomalies and possible risks \cite{ni2024timeseries}. By mining multi-dimensional data, LLM can assist personnel in recognizing abnormal patterns and provide an early alarm to support decision-makers reliably \cite{luo2025cross}.

\section{RELATED WORK}
Lakshmi et al. \cite{lakshmi2025proactive} use advanced machine learning techniques such as variational autoencoders and long short-term memory networks to monitor runtime behavior and detect anomalies such as zero-day attacks. Chauke et al. \cite{chaukeenhancing} propose an adaptive threat detection model designed to improve security in a multi-cloud environment. By applying machine learning and software-defined networking techniques, the model is able to analyze network traffic patterns, establish baselines of normal system behavior, and identify potential threats.

Madanan et al. \cite{madanan2024security} use machine learning and artificial intelligence techniques that are able to identify anomalous patterns in network traffic to identify potential security threats before they occur. Neol et al. \cite{neol2025ai} point to new requirements for current and future IT security measures for automated systems, proposing a variety of IT protection measures to address the current increasing threat landscape.

Kou et al. \cite{kou2024research} propose an improvement scheme for network security early warning system based on cloud computing platform. The solution adopts a multi-level data collection and analysis method, combined with machine learning and artificial intelligence technology, to achieve real-time monitoring and early warning of security incidents. Cai et al. \cite{cai2024multi} propose an intelligent multi-cloud resource scheduling system with endogenous security mechanism, which aims to improve the efficiency and security of resource scheduling in the cloud computing environment. In a multi-cloud architecture, information complexity, knowledge representation diversity, inference engine compatibility, and security threats in cloud networks make resource scheduling tasks extremely complex.

\section{METHODOLOGIES}
\subsection{Temporal and spatial features extraction}
In order to further extract deeper temporal and spatial features, we introduce a hybrid feature extraction method based on the combination of Deep Convolutional Network (CNN) and Recurrent Neural Network (RNN). This approach captures long-term dependencies and complex spatial changes in cloud environments more holistically. Suppose we have a multidimensional monitoring dataset $D \in \mathbb{R}^{N \times T \times m}$, where $N$ is the number of samples, $T$ is the time step, and $m$ is the characteristic dimension of each data point. First, we extract the spatial features through a convolution operation, which is defined as Equation (1):

\begin{equation}
F_{\text{CNN}} = Conv2D(D, W_1, b_1), \tag{1}
\end{equation}

where $W_1$ is the convolutional kernel and $b_1$ is the bias term, and the output is $F_{\text{CNN}} \in \mathbb{R}^{N \times T \times p}$, where $p$ is the eigendimension after convolution. Next, perform a non-linear activation operation, as shown in Equation (2):

\begin{equation}
F_{\text{CNN}}^{\text{ReLU}} = ReLU(F_{\text{CNN}} + b_1). \tag{2}
\end{equation}

To capture temporal features, we use a long short-term memory network, and recursive relationship is Equation (3):

\begin{equation}
h_t = \sigma(W_h x_t + U_h h_{t-1} + b_h), \tag{3}
\end{equation}

where $\sigma$ is the activation function, $W_h$ is the weight matrix of the LSTM, $x_t$ is the input feature of the time step $t$, and $h_t$ is the hidden state of the current time step.

In the proposed framework, the Large Language Model (LLM) serves as a high-level semantic encoder that supplements deep feature extraction. The input consists of log lines, condition strings, and event traces, which are pre-processed into structured sequences using template mining and keyword abstraction \cite{he2025givestructuredreasoninglarge, he2025selfgiveassociativethinkinglimited}. These sequences are tokenized and passed to the LLM, which outputs contextual embeddings $E_{\text{LLM}} \in \mathbb{R}^{n \times d}$, where $n$ is the number of tokens and $d$ is the embedding dimension.

By LSTM processing the features of each time step, the output $F_{\text{RNN}} \in \mathbb{R}^{N \times T \times q}$ is obtained, where $q$ is the output dimension of the LSTM. In order to better integrate contextual information, we combine LLM with traditional time series models and introduce a self-attention mechanism to enhance the context understanding ability of the model. Suppose the context vector generated by the LLM is $c_i$, where $c_i \in \mathbb{R}^k$, we further use the self-attention mechanism for information weighting. Assuming that the feature of the input is $z_i$, its weighted output $z_i^{att}$ is calculated by self-attention mechanism, as Equation (4):

\begin{equation}
z_i^{att} = \text{softmax}\left(\frac{z_i W_Q (z_i W_K)^\top}{\sqrt{d_k}}\right) \cdot z_i W_V, \tag{4}
\end{equation}

where $W_Q$, $W_K$ and $W_V$ are the weight matrices of the query, key, and value, respectively, and $d_k$ is the dimension of the key. By calculating the self-attention weights, the model is able to adjust the weights of features based on contextual information.

The CNN module adopts a 1D convolutional structure with three layers: kernel sizes of 3, 5, and 7, each followed by ReLU and batch normalization, capturing local correlations in metrics such as CPU usage, memory, and network I/O. The RNN module employs a 2-layer bidirectional LSTM with 128 hidden units per direction, processing time-ordered sequences of telemetry vectors with a window size of 10 steps. Feature outputs from the CNN (dimension 64) and LSTM (dimension 256) are concatenated at each timestamp.

\subsection{Multi-level feature extraction and anomaly detection}

After combining convolution and time series features and contextual information, we feed into a more complex anomaly detection model. We adopt a deep support vector machine (Deep SVM)-based approach to classify features extracted by deep neural networks. 

\begin{figure}[h!]
  \centering
  \begin{subfigure}[T]{1\linewidth}
  \includegraphics[width=1\linewidth, height=1\linewidth]{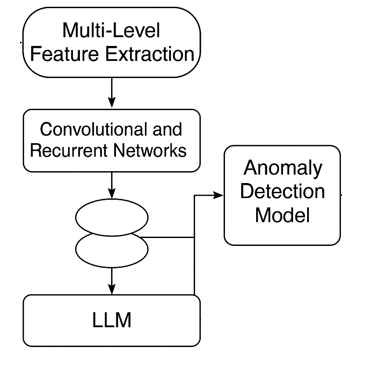}
  \end{subfigure}
  \caption{Architecture of Multi-level Feature Extraction and Anomaly Detection.}
  \label{fig1}
\end{figure}

Figure \ref{fig1} illustrates the overall architecture of multi-level feature extraction and anomaly detection. Firstly, the system extracts rich temporal series and spatial features from the input data through the multi-level feature extraction module. Then, the system uses a combination of convolutional neural network and recurrent neural network to further process these features to capture the spatial correlation and temporal dynamic behavior of the data. These processed features are then fed into an anomaly detection model that is used to identify potential anomalous patterns.

Assuming that our input features are $z_i^{att}$, the classification model is trained by minimizing the following loss function, as in Equations (5) and (6):

\begin{equation}
\mathcal{L}(w, b, \xi) = \frac{1}{2} \|w\|^2 + C \sum_{i=1}^{N} \xi_i + \sum_{i=1}^{N} L_{\text{hinge}} \left( y_i, f(z_i^{att}) \right), \tag{5}
\end{equation}

\begin{equation}
f(z_i^{att}) = w^\top z_i^{att} + b, \tag{6}
\end{equation}

where $f(z_i^{att})$ is the decision function of the support vector machine, $L_{\text{hinge}}$ is the hinge loss function, $\xi_i$ is the relaxation variable, and $C$ is the regularization parameter. We optimize this loss function by using gradient descent, updating the process to Equation (7):

\begin{equation}
w^{t+1} = w^t - \eta \nabla_w \mathcal{L}, \tag{7}
\end{equation}

where $\eta$ is the learning rate and $\nabla_w \mathcal{L}$ is the gradient of the loss function with respect to the weights.

We further introduce a complex early warning mechanism to quantify the uncertainty of the prediction results of the model by introducing Bayesian inference method. Assuming that the output of our anomaly detection model is $p(\hat{y}_i)$, we utilize Bayesian inference to calculate the confidence of the anomaly $p(\hat{y}_i \mid z_i)$, giving an early warning decision, as in Equation (8):

\begin{equation}
p(z_i) = \frac{p(\hat{y}_i) \, p(\hat{y}_i)}{p(z_i)}, \tag{8}
\end{equation}

where $p(\hat{y}_i)$ is the prior probability, $p(\hat{y}_i)$ is the likelihood function, and $p(z_i)$ is the marginal probability of the data. In this way, we are able to dynamically adjust the confidence level of the alert to ensure the reliability of the anomaly detection results.

The use of Deep SVM is motivated by its robustness in high-dimensional sparse feature spaces and its ability to capture nonlinear decision boundaries with limited training data. In this work, Deep SVM is implemented on top of the final feature layer using a radial basis function (RBF) kernel.

\section{EXPERIMENTS}
\subsection{Experimental setup}

The experiment used the Console Telemetry Dataset from the IBM Cloud platform. The dataset contains telemetry and log information from multiple microservices, covering more than 39,000 rows and 117,000 columns of data, which is highly dimensional, time-series, and dynamic, and is suitable for anomaly detection of large-scale cloud systems.

The performance of the proposed LLM-based model was comprehensively evaluated by comparing it with four thresholds-based anomaly detection methods including:

\begin{figure*}[!t]
  \centering
  \begin{subfigure}[T]{1\linewidth}    \includegraphics[width=1\linewidth, height=0.4\linewidth]{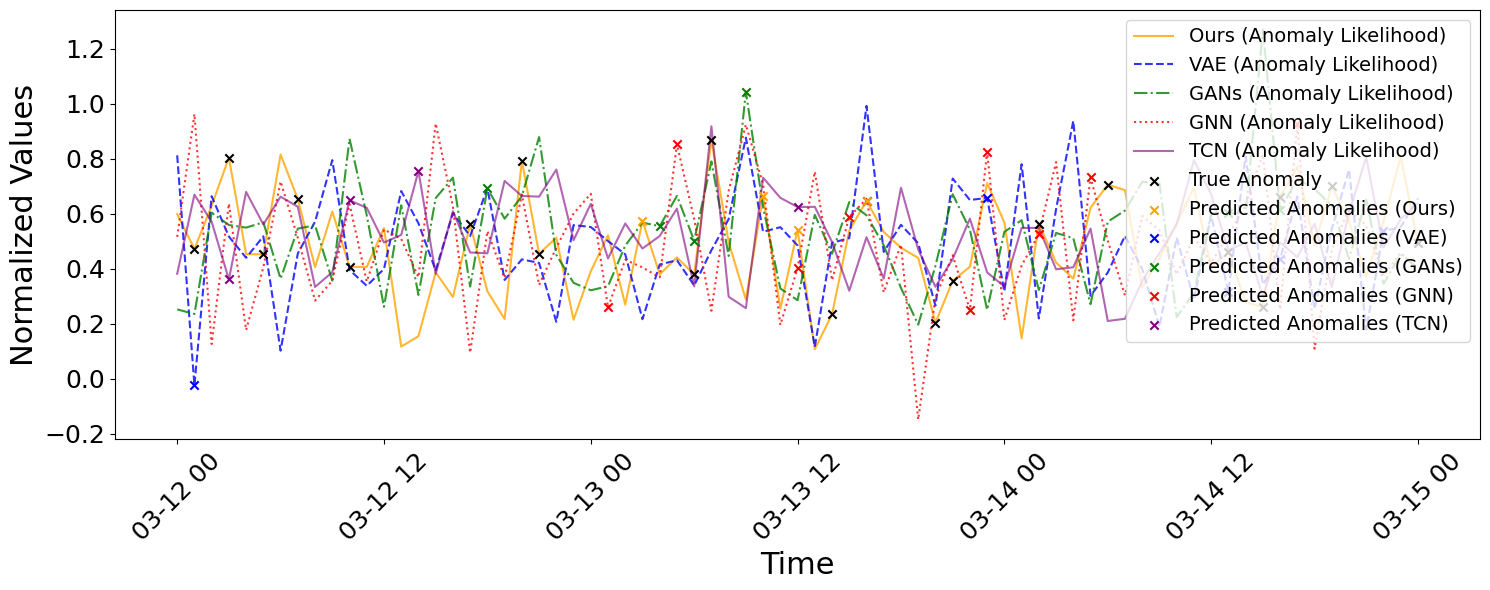}
  \end{subfigure}
  \caption{Anomaly Detection: Comparison of Methods.}
  \label{fig2}
\end{figure*}
\begin{itemize}[leftmargin=1.5em]

\item Variational Autoencoder (VAE) is a generative model that aims to learn an implicit representation of the data which maximizes the marginal likelihood of it \cite{kingma2013auto, wang2024fine, li2024exploring, li2024contextual}. In anomaly detection, VAE detects an anomaly by doing probabilistic modeling of the input data and through reconstruction error. 

\item Generative Adversarial Networks (GANs) are made of generator and discriminator that evolve simultaneously in a competition process through adversarial training to produce extremely realistic synthetic data \cite{goodfellow2020generative, li2024deception, yang2024comparative, xu2024style}. In the anomaly detection, GANs can be used to generate adversarial samples of an anomaly and measure similarities with real samples and find the patterns that characterize the anomaly. 

\item Graph neural networks (GNN) are deep learning networks for graph structured data \cite{zhou2020graph, liu2020handling, liu2023influence, yang2020revisiting, 2023arXiv231003272H, liu2024graphsnapshot, liu2025drtr}. GNN is able to capture GNN captures complex interactions between individual nodes and identifies abnormal behavior through information transfer mechanisms by modeling each service node in a cloud environment as a node. 

\item Temporal Convolutional Neural Networks (TCNs) through convolution operations capture long-term dependencies on the timeline, they have faster training speeds and more stable gradient propagation. In a multi-cloud context, TCN can learn quickly the time series behavior of a service from the cloud platform and can catch up the abnormal fluctuation of time.
\end{itemize}

\subsection{Experimental analysis}
Figure \ref{fig2} compares the performance of five anomaly detection methods (Ours, VAE, GANs, GNN, TCN) in a multi-cloud environment. Our model (Ours) successfully identified the vast majority of true anomalies over multiple time periods, and the predicted outliers were highly consistent with the true anomaly locations. In contrast, VAEs and GANs are able to detect anomalies in certain periods, but their predictions are less accurate and timely, especially during periods of high volatility, where false positives and false negatives occur.

\begin{figure}[h!]
  \centering
  \begin{subfigure}[T]{1\linewidth}
  \includegraphics[width=1\linewidth, height=0.5\linewidth]{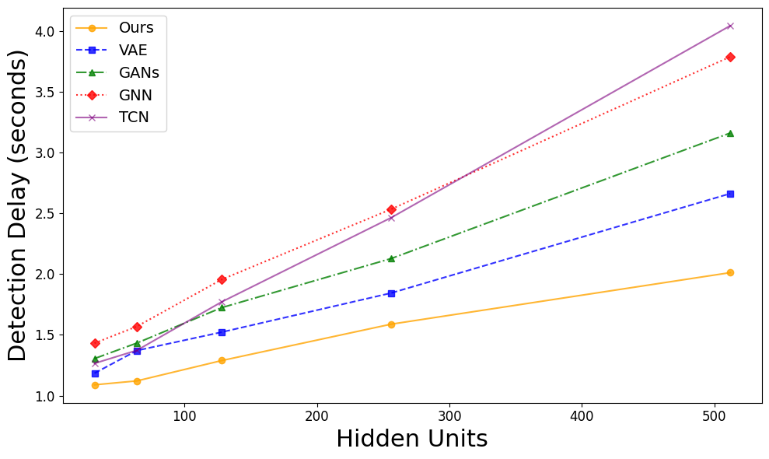}
  \end{subfigure}
  \caption{Comparison of Detection Delay Across Models.}
  \label{fig3}
\end{figure}

Response latency measures the time interval between when an anomaly occurs and when the model detects an anomaly. In Figure 3, we compare the detection latency of five anomaly detection methods at different hidden layer cell counts. With the increase of the number of hidden layer elements, the detection delay of all methods shows a gradual increasing trend, indicating that the complexity and computation of the model increase with the increase of network scale. In particular, GNN and TCN methods have a significant delay increase due to their complex network structures.

\section{CONCLUSION}
In conclusion, this work proposes an anomaly detection and early warning mechanism for intelligent monitoring system in multi-cloud environment based on Massive Language Model (LLM). By comparing with other advanced anomaly detection methods, the experimental results show that our method is superior to other methods in terms of accuracy and timeliness of anomaly detection, especially in the face of volatile and complex multi-cloud environments, and can more effectively capture sudden changes of anomaly. In addition, the detection delay analysis in the experiment shows that although the complexity and computational complexity of the model increase with the increase of the number of hidden layer elements, our model can balance the detection accuracy and response speed well, and demonstrates strong computational efficiency. Future research can further explore how to optimize the computational efficiency of LLM models in a multi-cloud environment and reduce the computational overhead of the model to adapt to larger-scale cloud platforms\cite{liu2025llmeasyquant, liu2024contemporary, liu2025fastcache}, and also investigate broader implications, such as ensuring fairness and mitigating potential biases in these AI-driven monitoring systems \cite{Ji2025}. Furthermore, a significant extension would be to move from anomaly detection to automated recovery, where integrating the LLM with a reinforcement learning framework could create an intelligent self-healing system, a promising direction supported by recent findings on LLMs in RL contexts \cite{zhao2024unveiling}.Finally, the optimization of our current model's training process itself could be enhanced by exploring more advanced algorithms inspired by recent progress in non-convex optimization \cite{xu2024stochastic, zhang2024agda+}.

\renewcommand{\bibfont}{\footnotesize}

\footnotesize{
\bibliographystyle{IEEEtran}
\bibliography{main}
}

\end{document}